\documentclass[acmtog, authorversion,nonacm]{acmart}

\usepackage{booktabs} 

\usepackage[ruled]{algorithm2e} 

\SetAlFnt{\small}
\SetAlCapFnt{\small}
\SetAlCapNameFnt{\small}
\SetAlCapHSkip{0pt} 

\usepackage{pifont}
\usepackage{xspace}
\usepackage{bm}

\usepackage{multirow}
\usepackage{siunitx}
\usepackage{booktabs}
\usepackage{adjustbox}
\usepackage{tabularx}
\usepackage{colortbl}
\usepackage{graphicx}
\usepackage[normalem]{ulem}
\usepackage{soul}

\newcommand{\cmark}{\textcolor{teal}{\ding{51}}}
\newcommand{\xmark}{\textcolor{red}{\ding{55}}}

\newcommand{\ie}{\textit{i.e.}\xspace}
\newcommand{\eg}{\textit{e.g.}\xspace}

\definecolor{best}{rgb}{0.96, 0.57, 0.58}
\definecolor{second}{rgb}{0.98, 0.78, 0.57}
\definecolor{third}{rgb}{1.0, 1.0, 0.56}

\begin{document}
\title{AnimaX: Animating the Inanimate in 3D with Joint Video-Pose Diffusion Models}

\author{Zehuan Huang}
\orcid{0009-0002-1883-0777}
\affiliation{%
 \institution{Beihang University}
 \country{China}}
\email{huangzehuan@buaa.edu.cn}

\author{Haoran Feng}
\affiliation{%
 \institution{Tsinghua University}
 \country{China}}
\email{fenghr24@mails.tsinghua.edu.cn}

\author{Yangtian Sun}
\affiliation{%
 \institution{The University of Hong Kong}
 \country{China}}
\email{sunyangtian98@gmail.com}

\author{Yuanchen Guo}
\authornote{Project leader.}
\affiliation{%
 \institution{VAST}
 \country{China}}
\email{imbennyguo@gmail.com}

\author{Yanpei Cao}
\authornote{Corresponding author.}
\affiliation{%
 \institution{VAST}
 \country{China}}
\email{caoyanpei@gmail.com}

\author{Lu Sheng}
\authornotemark[2]
\affiliation{%
 \institution{Beihang University}
 \country{China}}
\email{lsheng@buaa.edu.cn}

\begin{abstract}
We present AnimaX, a feed-forward 3D animation framework that bridges the motion priors of video diffusion models with the controllable structure of skeleton-based animation. Traditional motion synthesis methods are either restricted to fixed skeletal topologies or require costly optimization in high-dimensional deformation spaces. In contrast, AnimaX effectively transfers video-based motion knowledge to the 3D domain, supporting diverse articulated meshes with arbitrary skeletons. Our method represents 3D motion as multi-view, multi-frame 2D pose maps, and enables joint video-pose diffusion conditioned on template renderings and a textual motion prompt. We introduce shared positional encodings and modality-aware embeddings to ensure spatial-temporal alignment between video and pose sequences, effectively transferring video priors to motion generation task. The resulting multi-view pose sequences are triangulated into 3D joint positions and converted into mesh animation via inverse kinematics. Trained on a newly curated dataset of 160,000 rigged sequences, AnimaX achieves state-of-the-art results on VBench in generalization, motion fidelity, and efficiency, offering a scalable solution for category-agnostic 3D animation. Project page: \href{https://anima-x.github.io/}{https://anima-x.github.io/}.
\end{abstract}

%
%


%
%


\begin{teaserfigure}
  \includegraphics[width=\textwidth]{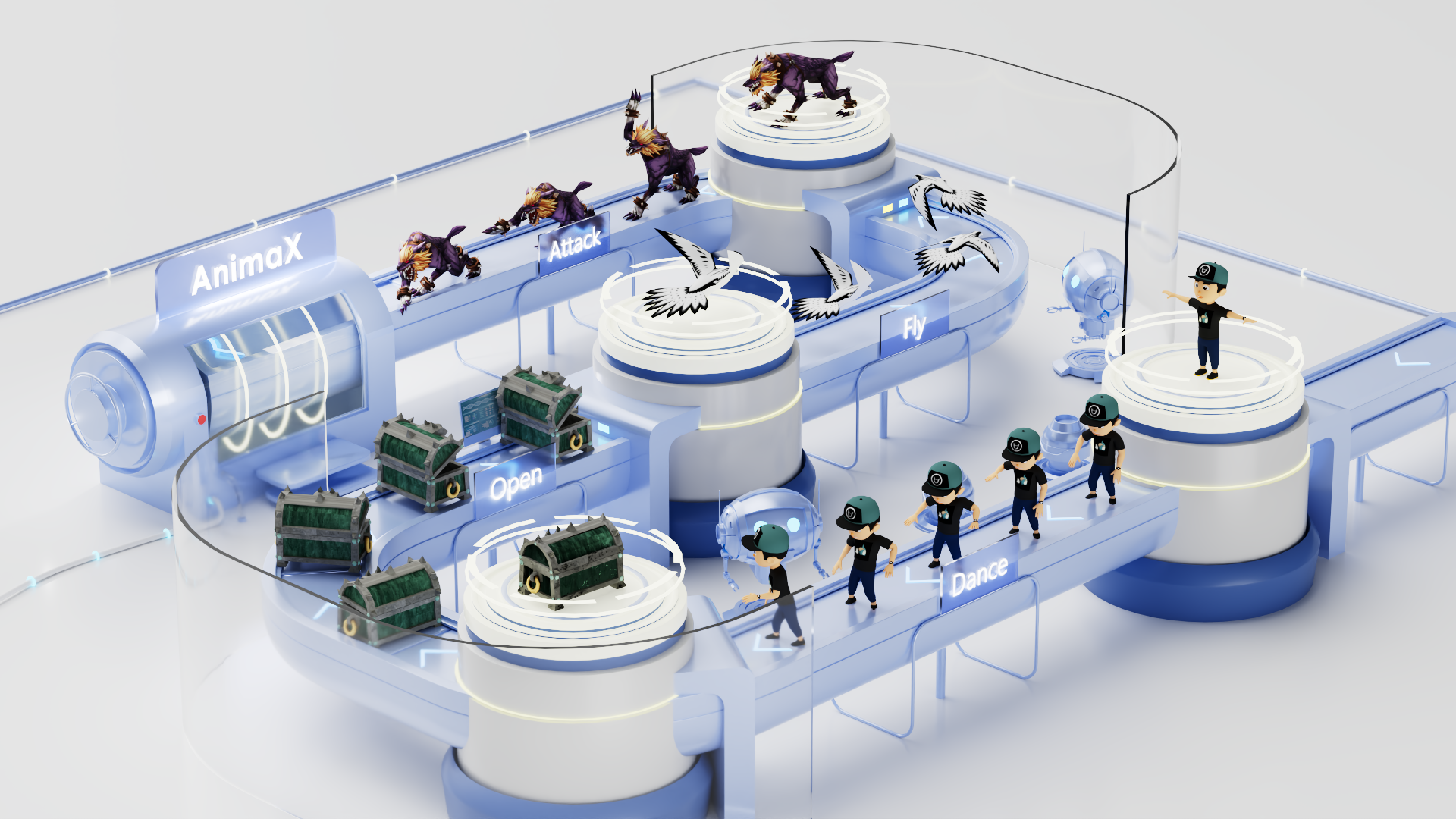}
  \caption{Diverse articulated 3D models animated using \textit{AnimaX}. The created animation, spanning various categories including humanoids, animals, and fictional models, demonstrates the versatility of our method. Selected models are visualized with keyframes of their predicted animations on the conveyor belts.}
  \label{fig:teaser}
\end{teaserfigure}

\maketitle

\begin{table*}
  \caption{Comparison of \textit{AnimaX} with existing generative work in 3D animation. Not all related methods are listed, but other approaches are generally similar to those included in the table.}
  \label{tab:comp_method}
  \centering
  \resizebox{1.0\linewidth}{!}{
  \begin{tabular}{lccccc}
    \toprule
    Method & Multi Categories & Skeleton-Based & Generative Prior & Output Format & Cost Time \\
    \midrule
    MotionDiffuse~\cite{motiondiffuse} & \xmark & \cmark & Motion Diffusion & Pose Sequence & 20 s   \\
    MDM~\cite{mdm} & \xmark & \cmark & Motion Diffusion & Pose Sequence & 25 s   \\
    ATU~\cite{atu} & \xmark & \cmark & Video Diffusion & Pose and Mesh Sequence & 1.5 hours  \\
    \midrule
    Diffusion4D~\cite{diffusion4d} & \cmark & \xmark & MV Video Diffusion & NeRF & 8 min \\
    Animate3D~\cite{animate3d} & \cmark & \xmark & MV Video Diffusion & GS & 45 min \\
    MotionDreamer~\cite{motiondreamer} & \cmark & \xmark & Video Diffusion & Mesh Sequence & 20 min \\
    \midrule
    AKD~\cite{li2025akd} & \cmark & \cmark & Score Distillation Sampling & GS & \textcolor{red}{25 hours} \\
    \textbf{AnimaX (Ours)} & \cmark & \cmark & MV Video-Pose Diffusion & Pose and Mesh Sequence & 6 min \\
    \bottomrule
  \end{tabular}
    }
\end{table*}

\section{Introduction}

In traditional computer graphics, skeleton-based character animation typically involve binding a skeleton to a mesh and defining keyframes for motions.
While this established technique affords high realism and fine-grained control over the resulting motion, it necessitates substantial manual effort from highly skilled artists, which is both time-consuming and expensive.

Recent advances in generative models~\cite{ho2020ddpm,peebles2023dit,radford2019language} offer promising avenues for automating the character animation pipeline.
Several studies have trained motion diffusion~\cite{motiondiffuse,mdm,mld} or auto-regressive models~\cite{t2m-gpt,motiongpt1} on collected motion capture data~\cite{amass,humanml3d}, enabling text-to-motion generation.
However, these models can only be trained on datasets with a pre-defined, fixed skeleton system (\ie, definition of joints with their connectivity), or rely on parametric 3D human models~\cite{loper2023smpl} for reconstruction.
These methods primarily support motion synthesis for a single skeletal topology, such as humanoid motion, limiting their ability to generate animations for more diverse character categories.

Another series of work~\cite{4dfy,dreamgaussian4d,diffusion4d,motiondreamer,animate3d} explores 3D animation by leveraging advanced video generation models~\cite{guo2024animatediff,blattmann2023svd,yang2024cogvideox,wang2025wan}, distilling their learned generalized dynamic motion into consistent 4D sequences.
As summarized in Tab.~\ref{tab:comp_method}, these methods commonly leverage multi-view video diffusion models~\cite{animate3d,diffusion4d} to guide the optimization of neural deformation fields~\cite{pumarola2021dnerf,wu20244dgs}, which predict displacements at each location within a 3D volume to deform a 3D shape.
The resulting animation is a temporal sequence of these deformed shapes.
While flexible, these approaches do not involve low-level skeleton-based motion representation, and instead introduce a large number of degrees of freedom (DoFs), making optimization challenging and often leading to in-consistent shapes and suboptimal quality.
More recently, AKD~\cite{li2025akd} distills articulated motion sequences from a pre-trained video diffusion model using Score Distillation Sampling (SDS)~\cite{poole2022dreamfusion}, simplifying the optimization by limiting the number of DoFs to that of a few joints.
But it requires expensive optimization that takes even 25 hours.

We focus on \uline{efficiently animating articulated 3D meshes with arbitrary skeletal structures} in a feed-forward manner, combining the diverse motion knowledge of video generation models~\cite{yang2024cogvideox,kong2024hunyuanvideo,wang2025wan} with the low-DoF control of skeleton-based animation.
Given an articulated mesh and a textual description, our goal is to generate 3D motion sequences that animate the mesh.
\dashuline{The challenge lies in encoding and decoding sparse 3D poses for diverse skeletal topologies.}
While representing motion as graphs is straightforward, it hinders leveraging motion priors within video generation models, which are fundamental for category-agnostic and motion-diverse 3D animation.

Our system, \textit{AnimaX}, addresses this by \uline{representing 3D motion as multi-view, multi-frame 2D pose maps, and adapting video diffusion models to generate such motion sequences in a feed-forward way}.
A straightforward baseline is to fine-tune a video diffusion model to generate pose sequences alone.
However, \dashuline{the sparsity of pose representation and modality gap between RGB and pose maps make fine-tuning challenging, and disrupts the learned spatial-temporal priors}, leading to distorted or nearly static pose outputs (Fig.~\ref{fig:ablation}).
To better preserve and transfer video-based motion priors, we reveal the spatial alignment between video frames and pose frames at each timestep, and introduce \textit{a joint video-pose diffusion model} that simultaneously predicts RGB videos and pose sequences.
Crucially, we apply shared positional encoding across corresponding tokens in both modalities, ensuring coherence between video and pose streams.
We find that this joint generation strategy—combined with shared positional encoding—allows the spatial-temporal priors learned from videos to be effectively grounded into the pose sequence generation process, resulting in more expressive motion outputs.
Finally, we reconstruct the 3D animation by triangulating joint positions from multi-view poses and applying inverse kinematics to compute the joint angles.

We trained our multi-view video-pose diffusion models on a newly curated dataset of nearly 160,000 rigged 3D animation sequences, encompassing diverse categories such as humanoids, animals, and furniture.
Evaluation on VBench~\cite{huang2023vbench} demonstrates that \textit{AnimaX} outperforms prior work in terms of generalizability across mesh categories, motion richness and naturalness, and efficiency.
Our contributions are summarized as follows:
\begin{itemize}
    \item We introduce AnimaX, an efficient feed-forward framework for animating diverse 3D articulated meshes with arbitrary skeletal structures. AnimaX uniquely bridges rich motion priors from video diffusion models with the controllability of skeleton-based animation, overcoming key limitations of prior fixed-topology or category-specific approaches.
    \item We represent 3D motion as multi-view, multi-frame pose maps and design a joint multi-view video-pose diffusion model that simultaneously generates videos and corresponding 2D pose map sequences. This model incorporates novel shared positional encodings and modality-specific embeddings to ensure robust spatio-temporal alignment between video and pose, enabling a highly effective transfer of motion knowledge from video models to 3D animation task.
    \item We contribute a new, large-scale dataset of approximately 160,000 rigged 3D animation sequences. This dataset, encompassing diverse categories (e.g., humanoids, animals, articulated objects), is crucial for training generalizable, category-agnostic animation models like AnimaX and will serve as a valuable resource for future research.
\end{itemize}

\section{Related Work}

\paragraph{Generative Models for 3D Animation.}
Recent advances in generative models~\cite{ho2020ddpm,peebles2023dit,esser2021taming,radford2019language} have spurred rapid progress in 3D animation.
A significant body of work focuses on category-specific motion generation, such as text-driven human motion synthesis~\cite{language2pose,temos,motionclip,motiondiffuse,mdm,mld,make-an-animation,omg,motion2to3,t2m-gpt,motiongpt1,motiongpt2,li2024lifting}.
For example, MDM~\cite{mdm} successfully employ diffusion models~\cite{ho2020ddpm} for this task, with subsequent work~\cite{mld} exploring latent diffusion models~\cite{rombach2022ldm}.
Others~\cite{motiongpt1,motiongpt2} leverage large language models~\cite{radford2019language,brown2020language} within the motion domain to support diverse motion-related tasks.
However, these models only adapt to a pre-defined, fixed skeletal structures, or rely on parametric 3D human models~\cite{smplify,loper2023smpl}, hindering the generation of diverse character animations.

\begin{figure*}
  \centering
  \includegraphics[width=0.95\linewidth]{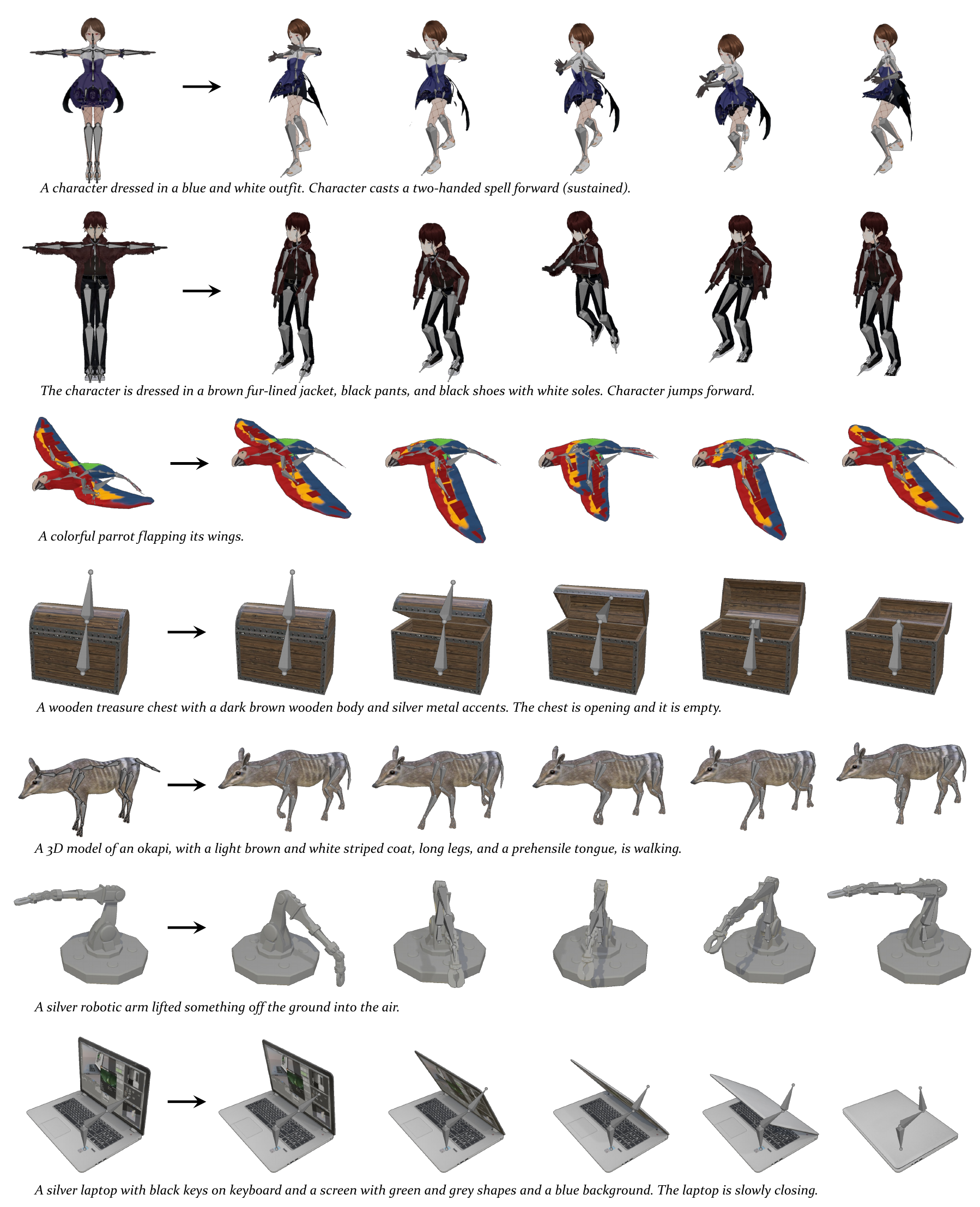}
  \caption{Animation results on generalized 3D models, including biped 3D assets, animals, chests, robotic arms.}
  \label{fig:results}
\end{figure*}

Another series of research~\cite{consistent4d,4dfy,alignyourgaussians,dreamgaussian4d,tc4d,stag4d,eg4d,diffusion2,diffusion4d,4diffusion,motiondreamer,ar4d,genxd,free4d,v2m4,shi2025driveanymesh,wu2025animateanymesh} explores 3D animation~\cite{write-an-animation,make-an-animation,zhang2024magicpose4d} by leveraging pre-trained image~\cite{rombach2022ldm,podell2023sdxl,flux2024}, video~\cite{ho2022vdm,makeavideo,guo2024animatediff,blattmann2023svd,bao2024vidu,yang2024cogvideox,kong2024hunyuanvideo,wang2025wan}, multi-view image~\cite{shi2023mvdream,liu2023syncdreamer,zuo2024videomv,huang2024epidiff,wen2024ouroboros3d,huang2024mvadapter,gao2024cat3d}, multi-view video~\cite{animate3d,diffusion4d,4diffusion,sv4d,yao2025sv4d2,bai2024syncammaster} diffusion models, distilling their generalized motion or 3D priors into 4D sequences.
Some approaches directly construct prior models, including diffusion~\cite{animate3d,diffusion4d,gat2025anytop} and reconstruction~\cite{l4gm} models, in the 4D domain.
Others distill 4D motion from a combination of generative models operating in lower dimensions, such as images, videos, and multi-view images.

Most relevant to \textit{AnimaX} are Diffusion4D~\cite{diffusion4d}, Animate3D~\cite{animate3d}, and MotionDreamer~\cite{motiondreamer}, which can generate animation from various 3D models.
Diffusion4D and Animate3D train multi-view video diffusion models on 4D data, and distill their spatial-temporal prior into 4D generation.
MotionDreamer extracts semantic motion priors from the deep features of video diffusion models to optimize deformation parameters in a zero-shot manner.
However, these methods do not explicitly represent or generate 4D content in a physically grounded way.

A more recent work, Articulated Kinematics Distillation (AKD)~\cite{li2025akd}, combines traditional skeleton-based character animation pipelines with generative models, introducing a more physically plausible approach.
Given a rigged 3D asset, AKD distills articulated motion sequences from video diffusion models using Score Distillation Sampling (SDS)~\cite{poole2022dreamfusion}.
However, AKD requires approximately 25 hours to optimize a single animation, motivating our focus on efficient, feed-forward articulated motion generation.

\paragraph{Video Diffusion Models.}
Video generation~\cite{ho2022vdm,blattmann2023alignyourlatents,chen2023videocrafter1,blattmann2023svd,yu2023animatezero,xing2024dynamicrafter} has developed rapidly in recent years.
Previous video diffusion models~\cite{guo2024animatediff} usually build upon image diffusion models~\cite{rombach2022ldm}, leveraging their pre-trained image knowledge by preserving spatial layers and inserting temporal layers to model motion dynamics.
These methods typically utilize image VAEs, failing to compress temporal information effectively, thus limiting generation to very short videos.
State-of-the-art video diffusion models~\cite{yang2024cogvideox,kong2024hunyuanvideo,wang2025wan,bao2024vidu} incorporate 3D causal VAEs, compressing both spatial and temporal dimensions of videos, coupled with diffusion transformers (DiTs)~\cite{peebles2023dit} that denoise in the latent space.
Trained on large-scale video datasets, these models demonstrated the capacity to generate diverse and realistic videos from textual prompts.
These implicitly learned motion priors provide a foundation for category-agnostic and motion-diverse 3D animation.

\begin{figure*}
  \centering
  \includegraphics[width=0.95\linewidth]{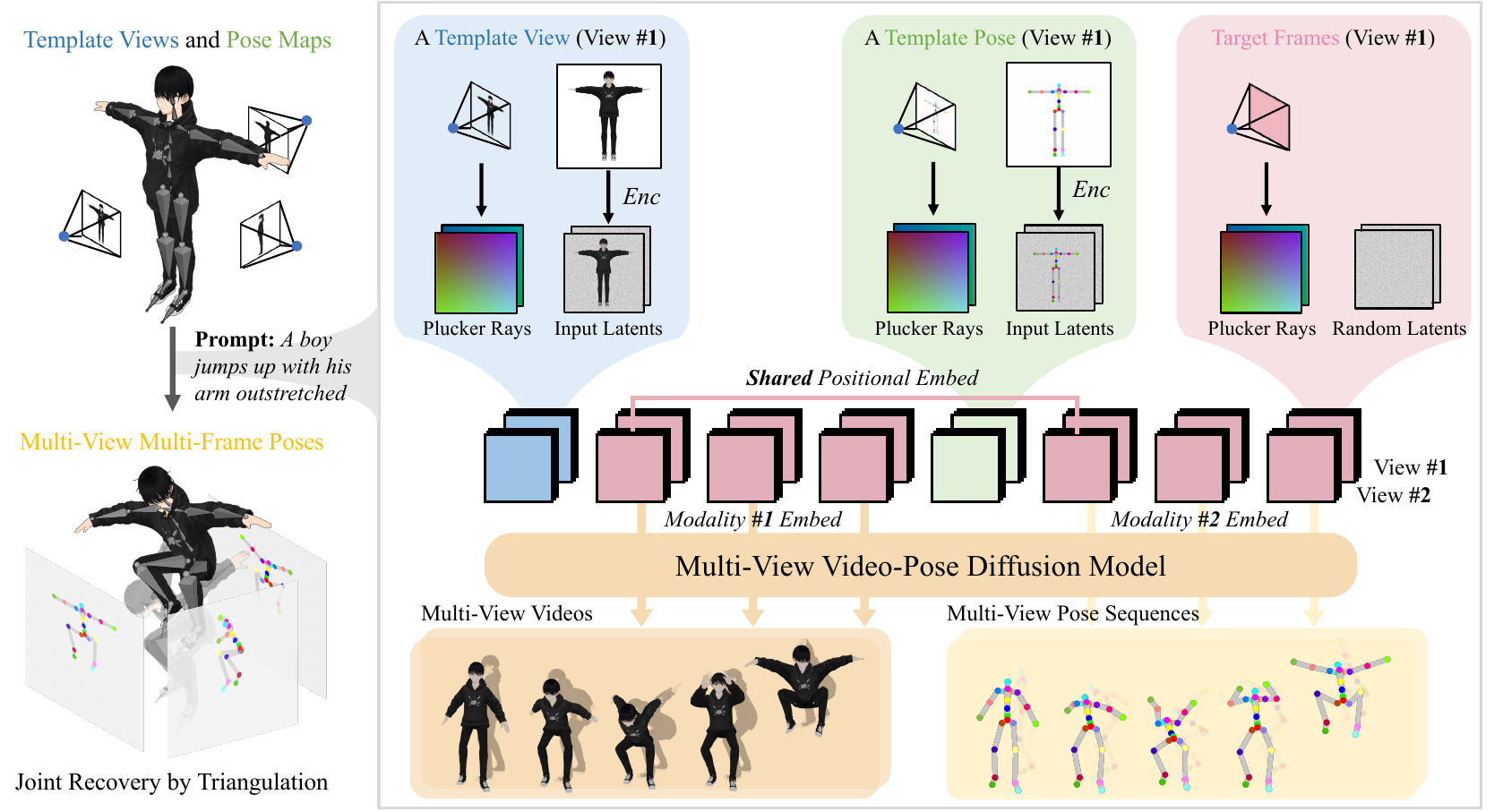}
  \caption{Illustration of \textit{AnimaX}. Given an articulated 3D mesh, \textit{AnimaX} creates a sequence of 3D animation in minutes. \textit{AnimaX} has two stages: (1) generating multi-view consistent videos and corresponding pose sequences simultaneously, conditioned on rendered template views and pose maps from the input mesh, with a textual description of the motion; and (2) recovering 3D joint positions per frame using multi-view triangulation~\cite{hartley1997triangulation} and applying inverse kinematics to obtain the joint angles and animate the mesh.}
  \label{fig:pipeline}
\end{figure*}

\section{Methodology}

\textit{AnimaX} generates 3D animations from a given articulated 3D mesh and a textual description of the motion.
\textit{AnimaX} has two stages: first, conditioned on the rendered views and pose maps from the mesh with a textual prompt, we generate multi-view consistent videos and corresponding pose sequences simultaneously using a joint video-pose diffusion model, and second, we reconstruct 3D motion from the generated multi-view pose sequences by multi-view triangulation and inverse kinematics (see Fig.~\ref{fig:pipeline}).
Below we describe our fundamental video diffusion models (Sec.~\ref{sec:vdm}), our multi-view video-pose diffusion model (Sec.~\ref{sec:mvvm}), and how the generated pose sequences are recovered to a 3D animation (Sec.~\ref{sec:recovery}).

\subsection{Preliminary: Video Diffusion Model}
\label{sec:vdm}

Video diffusion models~\cite{wang2025wan,kong2024hunyuanvideo} have demonstrated superior capabilities in generating generalized motion videos from textual descriptions, or animating arbitrary images guided by text.
Typically, these models include a 3D causal variational auto-encoder (VAE) and a diffusion transformer (DiT)~\cite{peebles2023dit} $\epsilon_{\theta}$ for denoising.
The VAE compresses the video's spatio-temporal dimensions; given a video $V \in \mathbb{R}^{(1+F)\times H\times W\times 3}$, the encodes maps it from pixel space to latent space $x\in \mathbb{R}^{(1+f)\times h\times w\times c}$.
A diffusion transformer is then trained in this latent space, progressively denoising latent variables into video latents, which the VAE decoder reconstructs back into video frames.

In diffusion transformer, positional information of video latents $x$ is typically encoded via RoPE~\cite{su2024rope}, applying rotation matrices based on each token's coordinate $(i,j,k)$ in a 3D grid:
\begin{equation}
    \dot{x}^{i,j,k} = x^{i,j,k} \cdot \bm{R}(i,j,k),
\end{equation}
where $\bm{R}(i,j,k)$ denotes the rotation matrix at position $(i,j,k)$ with $0\leq i < f$, $0\leq j<w$, and $0\leq k\leq h$.
Subsequently, 3D attention is applied to these position-encoded tokens to capture both intra-frame and inter-frame relationships, while cross-attention incorporates textual conditioning information $c^{txt}$.


\begin{figure*}
  \centering
  \includegraphics[width=\linewidth]{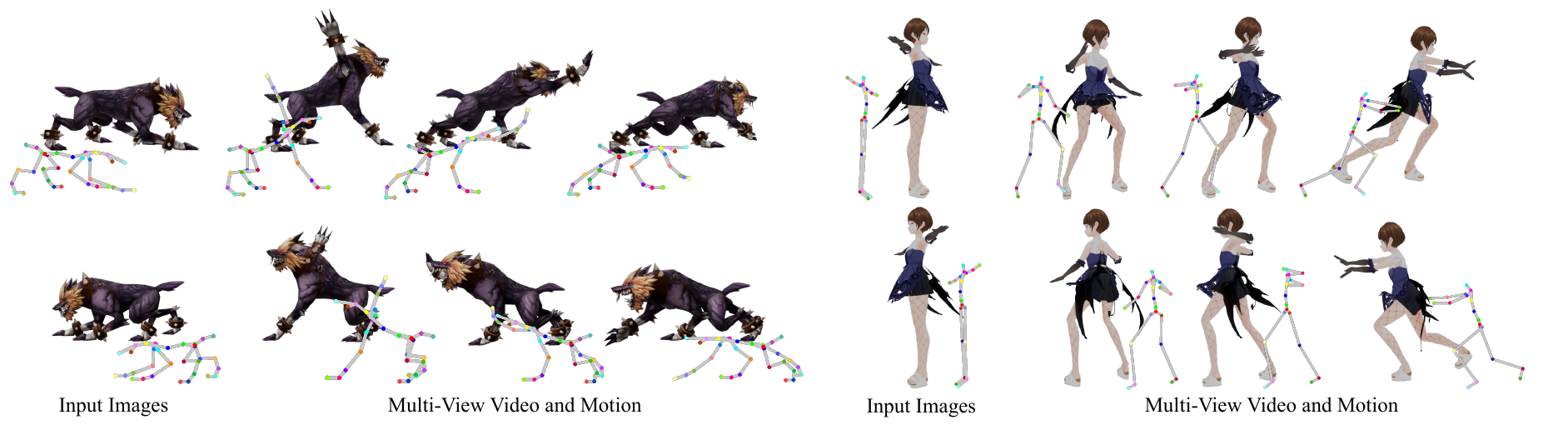}
  \caption{Input and output examples of our multi-view video-pose diffusion models, with input prompts ``A wolf is attacking something'' and ``A girl casting spell with hands forward''. We show two views here, but the model actually generates four views.}
  \label{fig:mvvm_examples}
\end{figure*}

\subsection{Multi-View Video-Pose Diffusion Model}
\label{sec:mvvm}

We train a multi-view video-pose diffusion model that takes multi-view images and pose maps of a static 3D articulated mesh as input, and generates multi-view videos and pose sequences given a textual motion description, as depicted in Fig.~\ref{fig:mvvm_examples}.
Specifically, given $N$ template views containing $N$ RGB images $I^{rgb}$, pose maps $I^{pose}$ and their corresponding camera parameters $C^{cam}$, the model learns to capture the joint distribution of $N$ view RGB videos $V^{rgb}$ and pose sequences $V^{pose}$ with the guidance of a textual prompt $C^{txt}$:
\begin{equation}
    p(V^{rgb}, V^{pose} | I^{rgb}, I^{pose}, C^{cam}, C^{txt})
\end{equation}

\paragraph{Pose Map Definition.}
We project the head positions of each bone in the skeletal animation onto the 2D image plane and assign a unique color to each joint for accurate localization during the subsequent recovery stage.
In the rendered image, joints are visualized as circular markers, while the skeletal structure is depicted by connecting lines between parent and child joints.

\paragraph{Model Architecture.}
Our model architecture is initialized from video latent diffusion models~\cite{wang2025wan}, but with additional camera and modality embeddings as well as enhanced positional embeddings for joint video-pose generation.
Given template RGB images, pose maps $I^{rgb}, I^{pose}\in \mathbb{R}^{H\times W\times3}$ and target RGB videos and pose sequences $V^{rgb}, V^{pose}\in \mathbb{R}^{(1+F)\times H\times W\times3}$ under $N$ views, the model encodes each image and video into a latent representation through a 3D causal VAE, to obtain image latent tokens $c^{rgb}, c^{pose}\in \mathbb{R}^{1\times h\times w\times c}$ and video latent tokens $x^{rgb}, x^{pose} \in \mathbb{R}^{(1+f)\times h\times w\times c}$ respectively.
Then, a diffusion transformer~\cite{peebles2023dit}, initialized from video DiT~\cite{wang2025wan}, is trained to estimate the joint distribution of the latent representations under multi-view capturing, given conditioning signals.

\paragraph{Image Conditioning and Temporal Modeling.}
We propose a simple yet unified approach that simultaneously enables image conditioning and consistent temporal modeling of two sequences of video latent tokens.
Given the input image tokens $c^{rgb}$, $c^{pose}$ and noisy video tokens $x_{t}^{rgb}$, $x_{t}^{pose}$, we concatenate them along the temporal dimension into a unified video token sequence $x^{total}\in \mathbb{R}^{(2f+4)\times h\times w\times c}$, where RGB video tokens are located in the first half while pose video tokens are in the second half.
We then extend the processing scope of existing 3D self-attention layers to jointly attend over the entire temporally-padded sequence:
\begin{equation}
    \text{Attention}([c^{rgb}, x_{t}^{rgb}, c^{pose}, x_{t}^{pose}]),
\end{equation}
where $c^{rgb}$ and $c^{pose}$ denote the clean, non-noised conditioning tokens, while $x_{t}^{rgb}$ and $x_{t}^{pose}$ are noisy latent tokens at timestep $t$.
This design leverages pre-trained 3D self-attention to seamlessly incorporate conditioning information from input images, efficiently utilizing the spatial-temporal priors of video diffusion models.
As a result, it enables encoding of visual details with strong compatibility to the generative framework.

\begin{figure*}
  \centering
  \includegraphics[width=\linewidth]{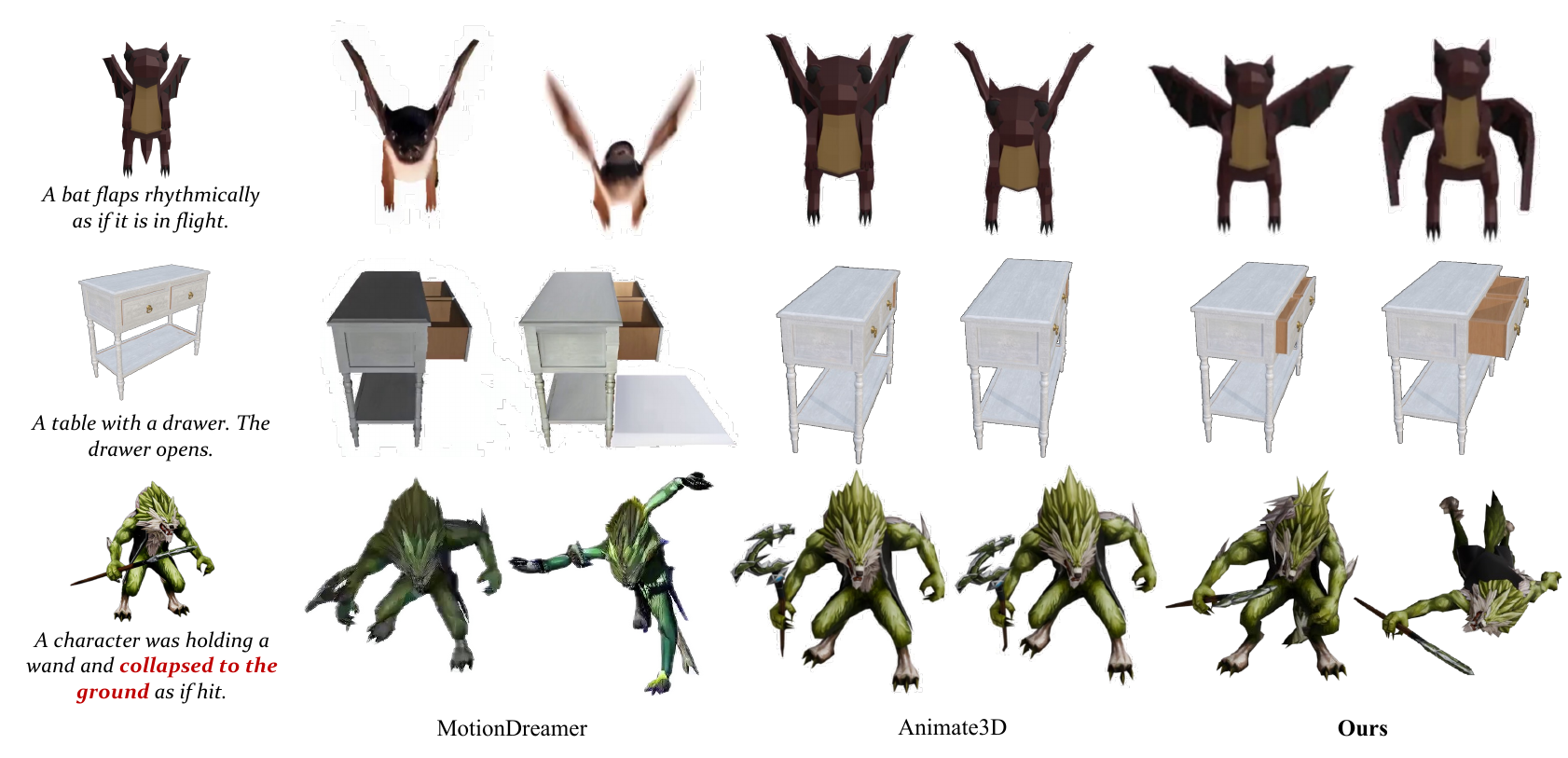}
  \caption{Comparison with state-of-the-art generalizable 4D generation methods. We compare our method with representative 3D-to-4D methods, including MotionDreamer~\cite{motiondreamer} and Animate3D~\cite{animate3d}. Our model can synthesize more correct and authentic animation clips compared to these methods which rely on optimization of neural deformation fields and do not involve low-level skeleton-based representation.}
  \label{fig:comparison}
\end{figure*}

\paragraph{Cross-Modal Modeling.}
To distinguish between the two modalities—RGB and pose—in the unified video token sequence $x^{total}$, we introduce an additional modality embedding.
We assign constant identifiers $0$ and $1$ to indicate the two modalities.
These identifiers are further transformed via frequency encoding, followed by several linear layers, yielding embeddings that share the same dimensionality as the original timestep embeddings.
The resulting embeddings are then added to the corresponding token's timestep embedding.

Furthermore, considering that in $x^{total}$, the first half $[c^{rgb}, x_{t}^{rgb}]$ and the second half $[c^{pose}, x_{t}^{pose}]$ are spatially aligned, we introduce a shared positional encoding mechanism to enforce structural consistency.
Specifically, for the sequence $x^{total}\in\mathbb{R}^{(2f+4)\times h\times w\times c}$, we define that tokens at positions $(i,j,k)$ and $(i+f+2, j, k)$ share the same positional encoding, formally expressed as:
\begin{equation}
\text{PE}^{i,j,k}=\text{PE}^{i+(f+2),j,k}=\bm{R}(i,j,k),
\end{equation}
where $\bm{R}(i,j,k)$ denotes the rotation matrix used in RoPE~\cite{su2024rope} at location $(i,j,k)$, with $0\leq i<f+2$, $0\leq j<w$, and $0\leq k<h$.
The design enables effective alignment and interaction between spatially corresponding tokens from different modalities.

\paragraph{Multi-View Consistency Modeling.}
To enable consistent multi-view video generation, we introduce additional camera conditioning and multi-view attention layers.
Specifically, we adopt Plücker ray map to represent camera poses~\cite{huang2024epidiff,sitzmann2021lightfield}.
For each view, the corresponding ray map is concatenated channel-wise to the latent representations of both the input images and the generated video frames.
To further enforce cross-view consistency, our multi-view layers operate on the multi-view video token sequence $x^{mv}\in\mathbb{R}^{N\times(2f+4)\times h\times w\times c}$, where $N$ denotes the number of views.
The token sequence is first inflated into $\dot{x}^{mv}\in\mathbb{R}^{(2f+4)\times(N\cdot h\cdot w)\times c}$, and self-attention is performed across the spatial dimension that aggregates all views.
This formulation enables the model to directly learn spatial correspondences and enforce consistency across different camera viewpoints.

\subsection{3D Motion Reconstruction and Animation}
\label{sec:recovery}
After obtaining multi-view pose sequences, we recover 3D poses and animate the 3D mesh through a three-stage process.
\textit{1) 2D joint localization:}
For each frame, we first extract 2D joint positions $p^{1:v}$ by clustering~\cite{arthur2006kmeans} the colors in the pose maps corresponding to each joint and taking the cluster centers as joint coordinates.
\textit{2) 3D joint optimization via triangulation:}
We then estimate the 3D joint positions $P^{1:v}$ by solving a non-linear least-squares optimization problem~\cite{hartley2003multiple}.
The objective is to minimize the re-projection error between the projected 3D joints and the observed multi-view 2D joint positions $\{p^{1:v}\}^{1:N}$, while enforcing bone length consistency.
\textit{3) Kinematic parameter estimation:}
Based on the joint positions in both the template pose and the predicted pose, we apply the inverse process of forward kinematics to estimate the animation parameters~\cite{aristidou2011fabrik}.
Traversing from the root node to the end-effectors, the rotation angle of each joint is estimated based on its positional deviation relative to the template pose.
The resulting joint angles are then applied to animate the articulated 3D mesh.

\section{Experiments}

We used Objaverse~\cite{deitke2023objaverse,deitke2023objaversexl}, Mixamo~\cite{mixamo}, VRoid~\cite{vroid} as our raw data source, and extracted a total of 161,023 animation clips after processing and cleaning.
Of these, we pick out 35 data pairs for evaluation, covering various categories such as humanoids, quadrupeds, birds, cabinets.
We render multi-view videos from the generated animation and use VBench~\cite{huang2023vbench} to evaluate on them.

We implement our multi-view video-pose diffusion model based on the Wan2.1~\cite{wang2025wan} text-to-video diffusion architecture, using the 1.3B parameter variant.
A two-stage training strategy is employed.
In the first stage, we fine-tune a single-view joint video-pose diffusion model using the LoRA~\cite{hu2021lora} technique to efficiently adapt the pretrained backbone.
In the second stage, we freeze all pretrained weights and train only the newly introduced camera embeddings and multi-view attention layers, thereby extending the model to support multi-view video-pose generation without disrupting the original learned priors.

\subsection{Main Results and Comparisons}

\begin{figure*}
  \centering
  \includegraphics[width=\linewidth]{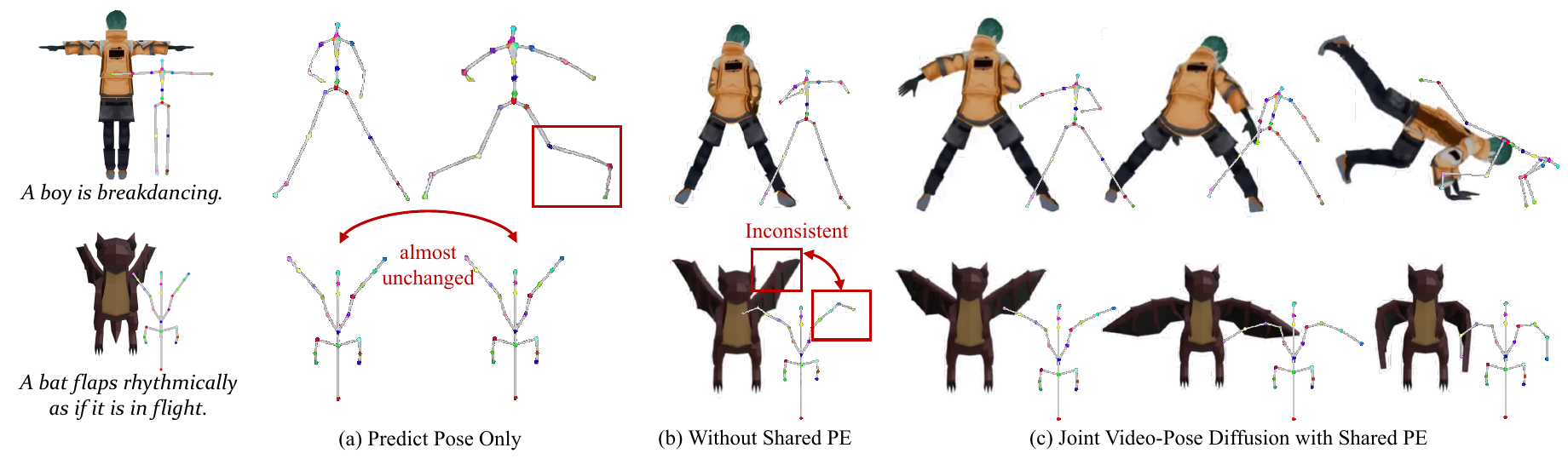}
  \caption{Qualitative ablation results on the joint video-pose diffusion model.}
  \label{fig:ablation}
\end{figure*}

\begin{table}[]
\centering
\caption{Quantitative comparisons. \textit{I2V Subject}, \textit{Smooth.}, \textit{Dynamic Deg.}, \textit{Quality} in VBench~\cite{huang2023vbench} are used to evaluate the consistency with the given image, the motion smoothness, the motion degree, and the appearance quality, respectively. Values of all metrics are the higher, the better, except for \textit{Dynamic Deg.}, since completely failed results (\eg, subject disappears) present an extremely high degree.}
\label{tab:comparison}
\resizebox{1.0\linewidth}{!}{
\begin{tabular}{l|cccc}
\toprule
Methods & I2V Subject$\uparrow$ & Smooth.$\uparrow$ & Dynamic Deg. & Quality$\uparrow$ \\
\midrule
Animate3D & 0.943 & 0.986 & 0.446 & 0.481 \\
MotionDreamer & 0.817 & 0.977 & 0.827 & 0.439 \\
Ours & \textbf{0.962} & \textbf{0.990} & 0.661 & \textbf{0.517} \\
\bottomrule
\end{tabular}
}
\end{table}

\begin{table}[]
\centering
\caption{User study results on 3D animation. We collected user preference on motion-text alignment, 3D shape consistency, and overall motion quality from 30 participants. Our method receives the best preference on all metrics.}
\label{tab:user_study}
\resizebox{1.0\linewidth}{!}{
\begin{tabular}{l|rrr}
\toprule
Methods & Motion-Text Align.$\uparrow$ & Shape Consist.$\uparrow$ & Overall Motion.$\uparrow$ \\
\midrule
Animate3D & 12.8\% & 26.7\% & 19.2\% \\
MotionDreamer & 4.3\% & 0.0\% & 2.9\% \\
Ours & \textbf{82.9\%} & \textbf{73.3\%} & \textbf{77.9\%} \\
\bottomrule
\end{tabular}
}
\end{table}

We present our primary results, which include the animated 3D conditioned on a 3D mesh and a text prompt, as illustrated in Fig.~\ref{fig:teaser}.
In Fig.~\ref{fig:mvvm_examples}, we display examples of our multi-view video-pose diffusion models.
Our results demonstrate that the model, adapted from pre-trained video diffusion models, jointly generates consistent and high-fidelity videos and motion sequences.

\paragraph{Qualitative Comparisons.}
We perform a qualitative comparison with Animate3D~\cite{animate3d} and MotionDreamer~\cite{motiondreamer}, two representative 3D-to-4D methods that utilize multi-view video diffusion models to guide the optimization of neural deformation fields.
As shown in Fig.~\ref{fig:comparison}, both baselines exhibit notable limitations.
MotionDreamer relies on pretrained video diffusion models to supervise deformation optimization; however, the excessive degrees of freedom (DoFs) inherent in deformation fields lead to inconsistent geometry and unstable temporal behavior.
Animate3D fine-tunes a multi-view video diffusion model to improve cross-view consistency.
While it reduces artifacts, the reconstruction remains challenging and often results in near-static outputs.
In contrast, our method transfers video-based motion priors into the motion synthesis task via joint video-pose modeling.
This enables the generation of temporally coherent and semantically aligned 3D animations, demonstrating superior consistency and motion expressiveness without requiring costly optimization procedures.

\paragraph{Quantitative Comparisons.}
Following \cite{animate3d}, we render multi-view videos from animated 3D and evaluate them with VBench~\cite{huang2023vbench}, a comprehensive video evaluation toolkit.
We choose 4 image-to-video metrics, \ie, \textit{I2V Subject}, \textit{Motion Smoothness}, \textit{Dynamic Degree}, and \textit{Aesthetic Quality}, measuring the consistency with the given image, the motion smoothness, the motion degree, and the appearance quality, respectively.
Values of all metrics are the higher, the better.
The \textit{Dynamic Deg.} metric aims to capture motion richness; however, it can be less robust, as severe generation failures (e.g., subject disappearing) may also produce misleadingly high scores.
As shown in Tab.~\ref{tab:comparison}, our method outperforms other methods, especially in appearance quality, due to our low-DoF animation design.

\paragraph{User Study.}
We conducted a user study comparing our approach with baseline 3D-to-4D methods~\cite{animate3d,motiondreamer}.
The study aimed to evaluate motion-text alignment, 3D shape consistency, and overall motion quality.
A total of 30 participants were recruited to provide their preferences between the outputs of difference methods on test set.
As shown in Tab.~\ref{tab:user_study}, our method receives the best preference on all metrics.
This highlights the superior capability of our method in transferring the motion priors of video diffusion models to skeleton-based 3D animation.

\subsection{Ablation Studies}

\begin{table}[]
\centering
\caption{Quantitative ablation results on the video-pose diffusin model design. Based on the video diffusion model, we fine-tune a model that (a) only predicts pose sequences and (b) a joint video-pose diffusion model without shared positional encoding mechanism, and (c) our full setting. Vbench~\cite{huang2023vbench} is used to evaluate on multi-view renderings.}
\label{tab:ablation}
\resizebox{1.0\linewidth}{!}{
\begin{tabular}{l|rrrr}
\toprule
Methods & I2V Subject$\uparrow$ & Smooth.$\uparrow$ & Dynamic Deg. & Quality$\uparrow$ \\
\midrule
(a) Pose Only & 0.960 & 0.982 & 0.402 & 0.448 \\
(b) w/o Shared PE & 0.954 & 0.988 & 0.660 & 0.429 \\
(c) Full Setting & \textbf{0.962} & \textbf{0.990} & 0.661 & \textbf{0.517} \\
\bottomrule
\end{tabular}
}
\end{table}

We ablate the key design in our joint video-pose diffusion models.
Specifically, based on video diffusion models, we examine three settings:
(a) a pose diffusion model that is fine-tuned to generate pose sequences alone,
(b) a video-pose diffusion model that generates videos and pose sequences simultaneously but do not share the same positional encoding in these two modalities,
and (c) our full video-pose diffusion model with shared positional encoding mechanism.

As illustrated in Fig.~\ref{fig:ablation} (a), directly fine-tuning a pre-trained video diffusion model to generate only pose sequences often leads to degraded performance due to the sparsity of pose representation and the significant modality gap between RGB videos and pose maps, as well as the relatively sparse supervision available for the pose modality.
This mismatch disrupts the learned spatial-temporal priors of the original model, frequently resulting in degenerate outputs such as distorted pose frames or nearly static pose sequences.
Therefore, we adopt a joint video-pose diffusion framework that simultaneously generates both RGB video frames and pose sequences.
Within this setting, we observe that compared to a baseline model (b), our full model—which shares positional encodings across the two modalities—significantly improves the spatial alignment between generated pose sequences and RGB videos.
This architectural design \uline{facilitates more effective transfer of pre-trained video priors to the motion generation task}, resulting in more coherent and realistic pose outputs.
We also lifted the pose sequences generated by these three models to 3D animation clips, and used VBench~\cite{huang2023vbench} to evaluate on them.
As shown in Tab.~\ref{tab:ablation}, model (c) performs the best, confirming the effectiveness of our model design.

\section{Conclusion}

We present AnimaX, a feed-forward framework for animating articulated 3D meshes with arbitrary skeletal structures by bridging the generalizable motion priors of video diffusion models with the structured controllability of skeleton-based animation.
Unlike prior approaches that either rely on fixed skeletal topologies or require costly optimization, our method enables efficient generation of temporally and spatially consistent multi-view pose and video sequences conditioned on a textual motion prompt.
By introducing joint video-pose diffusion, shared positional encodings, and modality-aware embeddings, AnimaX effectively transfers video-based motion knowledge to the 3D domain and supports a broad spectrum of mesh categories.
Extensive experiments on VBench validate the superiority of our method in terms of generalization, animation quality, and runtime efficiency.
We believe this work opens new avenues for scalable, category-agnostic 3D animation driven by text and visual priors.

\bibliographystyle{ACM-Reference-Format}
\bibliography{sample-bibliography}

\appendix
\section{Appendix}

\subsection{Dataset}

\paragraph{Data Curation.}
We use Objaverse~\cite{deitke2023objaverse,deitke2023objaversexl}, Mixamo~\cite{mixamo}, VRoid~\cite{vroid} as our raw data source.
For Objaverse and Objaverse-XL, we curate high-quality animation clips by applying the following filtering criteria: 1) The asset must contain both rigging and animation data. 2) Geometry and texture quality meet a fidelity threshold. 3) The animation sequence must contain more than 16 frames. 4) We discard clips with negligible motion based on the mean optical flow magnitude.
After filtering, we obtain a total of 48,020 animation clips from Objaverse and Objaverse-XL.
For Mixamo and VRoid, we retarget motion-only animation clips from Mixamo humanoid characters onto both Mixamo and VRoid static characters.
We then apply the same filtering strategy as used for Objaverse.
This process yields 55,222 animation clips from Mixamo characters and 57,781 stylized animation clips from VRoid anime-style characters.

In total, we collect 161,023 high-quality animation clips across multiple categories.
For evaluation, we sample 35 representative sequences covering a wide range of object categories, including humanoids, quadrupeds, flying animals, and articulated furniture.

\paragraph{Category Annotation.}
Following ~\cite{song2025magicarticulate}, we render each 3D model from four predefined viewpoints and arrange the resulting images into a 2×2 grid.
We then utilize GPT-4o~\cite{gpt4o} to perform automatic category labeling based on the composed images.

\begin{table*}[]
\small
\centering
\caption{Animation counts for each category in the curated dataset. Each column is sorted in descending order of animation count.}
\label{tab:my_label}
\resizebox{1.0\linewidth}{!}{
\begin{tabular}{lc|lc|lc|lc}
\toprule
Category & \# Animation & Category & \# Animation & Category & \# Animation & Category & \# Animation \\
\midrule
character& 140,000 & sculpture &1132  & furniture & 487 & miscellaneous & 196 \\
anthropomorphic &  22881 & vehicle & 1085  & scanned data & 425  & planet & 191 \\
toy & 12725  & anatomy & 1070  & electronic device & 410 & sporting goods & 140 \\
animal & 8603  & household item & 679  &  architecture &378 & paper & 77 \\
mythical creature & 5428 & plant & 606  & clothing & 304  & jewelry & 62 \\
weapon & 3221  & accessory & 570 &  food & 270 &  musical instrument & 31  \\
tool & 1297  & &  & & & & \\
\bottomrule
\end{tabular}
}
\end{table*}

\paragraph{Rendering.}

To support multi-view supervision, we render four-view videos and their corresponding poses from animation clips and their models in rest pose using Blender scripts~\cite{bpyrenderer}.
The camera setup consists of a fixed elevation angle of 0°, with azimuth angles set to 0°, 90°, 190°, and 270°, respectively.
Additionally, we generate video captions using the vision-language model Qwen2.5-VL~\cite{Qwen2.5-VL}.
These paired multi-view videos and pose maps form the training data for our multi-view video-pose diffusion model.

\subsection{Implementation Details}

\paragraph{Training.}
We implement our multi-view video-pose diffusion model based on the Wan2.1~\cite{wang2025wan} text-to-video diffusion architecture, using the 1.3B parameter variant and diffusers codebase~\cite{von-platen-etal-2022-diffusers}.
A two-stage training strategy is employed.
In the first stage, we fine-tune a single-view joint video-pose diffusion model using the LoRA~\cite{hu2021lora} technique to efficiently adapt the pretrained backbone.
In the second stage, we freeze all pretrained weights and train only the newly introduced camera embeddings and multi-view attention layers, thereby extending the model to support multi-view video-pose generation without disrupting the original learned priors.

The video-pose diffusion model is trained to generate videos at a resolution of 480p, with a sequence length of up to 81 frames.
\uline{Since Mixamo and VRoid datasets primarily consist of humanoid characters, we assign sampling weights during training to balance contributions from different data sources.}
Specifically, training samples are drawn from Mixamo and VRoid with probabilities of 0.25 each, and from Objaverse~\cite{deitke2023objaverse,deitke2023objaversexl} with a probability of 0.5.
We randomly selected either a frame from the video or a rendered view of the mesh in its rest pose as the image condition. This image condition was dropped with a probability of $0.2$.
We train the model for 5 epochs using a learning rate of $5\times 10^{-5}$ on 16 NVIDIA A100 GPUs.

\paragraph{Inference.}
At inference time, given an articulated 3D mesh, we first render four-view images and corresponding pose maps as templates. These are fed into our multi-view video-pose diffusion model, along with a textual motion description. For the denoising process, we set the image condition guidance scale to $3.0$, and use $50$ denoising steps. The model simultaneously generates four-view videos and pose sequences in approximately 5 minutes.
Subsequently, we extract 2D joint positions from the generated pose maps and lift them to 3D joint angles via multi-view triangulation followed by kinematic parameter estimation.
The total inference time, including both video generation and 3D reconstruction, is approximately 6 minutes.

\subsection{Limitations and Discussions}
Currently, our video-pose diffusion model generates videos from a fixed set of camera viewpoints with limited and static fields of view, making it challenging to synthesize animations with large spatial motion. However, we believe that our network architecture could potentially handle such scenarios if trained on videos captured with more flexible and dynamic camera movements. Under this setting, it would be possible to specify arbitrary camera trajectories during inference, enabling the generation of multi-view videos that capture wider motion ranges and larger 3D spaces.

In addition, our model inherits the limitation of pretrained video diffusion backbones, which typically constrain the maximum video length. As a result, generating temporally continuous long-form animations remains difficult. We anticipate that incorporating test-time training~\cite{dalal2025videottt} or autoregressive denoising~\cite{chen2024diffusionforcing} extensions may offer a promising direction to support the generation of longer and more coherent video sequences.


\end{document}